\documentclass[runningheads]{llncs}
\usepackage{graphicx}

\usepackage{tikz}
\usepackage{comment}
\usepackage{amsmath,amssymb} 
\usepackage{color}
\usepackage{graphicx}
\usepackage{comment}
\usepackage{epsfig}
\usepackage{algpseudocode,algorithm,algorithmicx}
\usepackage{url}            
\usepackage{booktabs}       
\usepackage{amsfonts}       
\usepackage{nicefrac}       
\usepackage{microtype}      
\usepackage{graphicx}
\usepackage{mathrsfs,amsmath,amssymb,mathtools}
\usepackage{multirow}
\usepackage{bm}
\usepackage{tabularx}       
\usepackage{float}
\usepackage{array}

\usepackage[width=122mm,left=12mm,paperwidth=146mm,height=193mm,top=12mm,paperheight=217mm]{geometry}

\newcommand{\C}{\mathbf{C}}
\newcommand{\D}{\mathbf{D}}

\newcommand{\M}{\mathbf{M}}

\newcommand{\bS}{\mathbf{S}}
\newcommand{\T}{\mathbf{T}}

\newcommand{\X}{\mathbf{X}}

\newcommand{\m}{\mathbf{m}}

\newcommand{\bt}{\mathbf{t}}

\newcommand{\x}{\mathbf{x}}

\newcommand{\real}{\mathbb{R}}
\newcommand{\bina}{\{0, 1\}}

\newcommand{\udots}{\mathinner{\mskip1mu\raise1pt\vbox{\kern7pt\hbox{.}}\mskip2mu\raise4pt\hbox{.}\mskip2mu\raise7pt\hbox{.}\mskip1mu}}

\DeclareMathOperator*{\argmax}{arg\,max}

\newenvironment{aligns}{\par\nobreak\small\noindent\align}{\endalign}

\definecolor{dark}{rgb}{0.15294118, 0.54117647, 0.76862742}
\definecolor{cyan}{rgb}{0.40392157, 0.61568627, 0.65882353}	
\definecolor{ncs}{rgb}{0.0, 0.53, 0.74}

\usepackage[pagebackref=false,
            breaklinks=true,
            colorlinks,
            bookmarks=false,
            hypertexnames=false]{hyperref}
\hypersetup{citecolor=black}

\begin{document}
\pagestyle{headings}
\mainmatter
\def\ECCVSubNumber{2339}  

\title{Matching Guided Distillation} 

\titlerunning{Matching Guided Distillation}
%
\author{Kaiyu Yue \and
Jiangfan Deng \and
Feng Zhou}
\authorrunning{K. Yue et al.}
%
\institute{Algorithm Research, Aibee Inc.}
\maketitle

\begin{abstract}
Feature distillation is an effective way to improve the performance for a smaller student model, which has fewer parameters and lower computation cost compared to the larger teacher model.
Unfortunately, there is a common obstacle \textemdash \  the gap in semantic feature structure between the intermediate features of teacher and student.
The classic scheme prefers to transform intermediate features by adding the adaptation module, such as naive convolutional, attention-based or more complicated one.
However, this introduces two problems:
a) The adaptation module brings more parameters into training.
b) The adaptation module with random initialization or special transformation isn't friendly for distilling a pre-trained student.
In this paper, we present Matching Guided Distillation (MGD) as an efficient and parameter-free manner to solve these problems.
The key idea of MGD is to pose matching the teacher channels with students' as an assignment problem.
We compare three solutions of the assignment problem to reduce channels from teacher features with partial distillation loss.
The overall training takes a coordinate-descent approach between two optimization objects \textemdash \ assignments update and parameters update.
Since MGD only contains normalization or pooling operations with negligible computation cost, it is flexible to plug into network with other distillation methods.
The project site is \textcolor{dark}{\url{http://kaiyuyue.com/mgd}}.
\end{abstract}
\section{Introduction}
\label{sec:intro}
\begin{figure}[t]
  \centering
  \includegraphics[width=1.\linewidth]{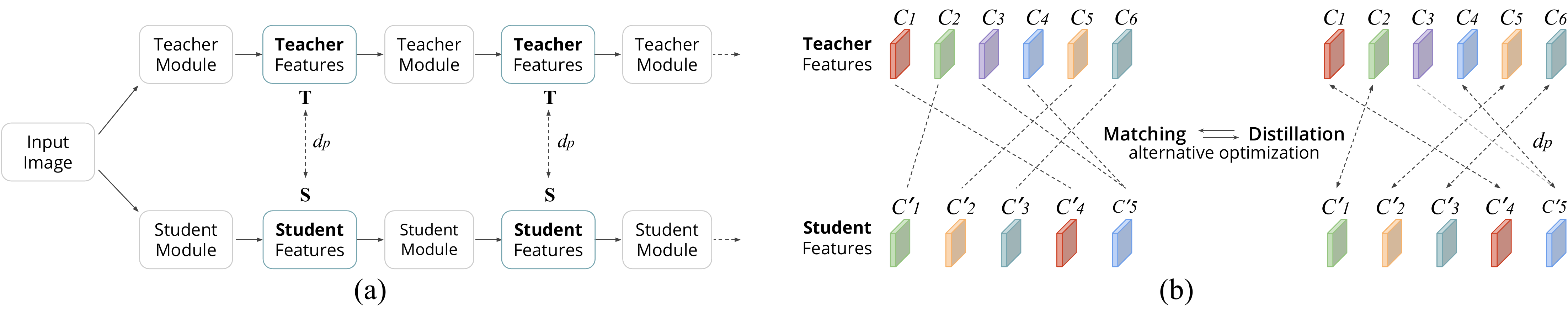}
  \caption{\small{
    \textbf{Matching guided distillation}. 
    (a) MGD follows the general distillation paradigm.
    $\T$ and $\bS$ are teacher and student feature tensors. 
    $d_p$ is the feature distance function.
    (b) MGD distills teacher channels (${C_i}$) of intermediate features to student channels ($C'_{i}$) by solving a joint matching-distillation problem (alternative optimization).
  }}
\label{fig:intro}
\end{figure}
Deep networks~\cite{zagoruyko2016wide,krizhevsky2012imagenet} enjoy massive neuron parameters for achieving the state-of-the-art performances on lots of technique lines, such as visual recognition~\cite{he2016deep}, image captioning~\cite{densecap}, object detection system~\cite{wu2019detectron2} and language understanding~\cite{yang2019xlnet,devlin2018bert}.
However, the industry prefers to carry out model inference on cheap devices, therefore the small model with few parameters is needed.
The dilemma of achieving analogous performance to the large model in lightweight backbone recently motivates extensive research directions, such as channel pruning~\cite{he2017channel}, lightweight model design~\cite{ma2018shufflenet,sandler2018mobilenetv2}, quantization~\cite{wang2019haq} and neural architecture search (NAS) for efficient model~\cite{wu2019fbnet,tan2019efficientnet}.
Among them, model distillation is another active track, which aims to transfer knowledge or semantic feature information from a large teacher model into a light student model.
Pioneered by dark knowledge~\cite{hinton2015distilling}, the main body of recent works~\cite{yim2017gift,chen2017learning} focuses on using distilling intermediate features to enrich the learnable information for guiding student in different tasks, such as classification~\cite{romero2014fitnets} and detection~\cite{chen2017learning}.
 
However, a prominent challenging problem for these methods is on how to fill the gap in semantic feature structure between teacher and student.
Roughly speaking, the contrast lies in two aspects:
1) The different channel dimensions of feature outputs used for distillation.
2) The different perceptual information or activations between two channel sets.
Previous works~\cite{romero2014fitnets,yim2017gift,heo2019overhaul} overcome these two obstacles by building an adaptive module between hidden layers of teacher and student.
Whereas this manner can alleviate the issue, it still has two limitations:
1) The adaptation module introduces more parameters (including weights, gradients and optimizer states) into training~\cite{pudipeddi2020training}.
These additional parameters induce the training harder despite the fact that they would be brushed off for model inference.
2) The adaptation module with random initialization or special non-parameter transformation~\cite{srinivas2018knowledge,Zagoruyko2017AT} isn't friendly for distilling a pre-trained student, because it would potentially disturb the student features.
To avoid this break, \cite{yim2017gift,romero2014fitnets} performs stage-wise training by separating optimization into multiple steps.
But this way isn't perfect yet because it will plunge training into a cumbersome one.

To crack the limitations of former works, we propose a novel distillation method named Matching Guided Distillation (MGD).
As shown in Fig.~\ref{fig:intro}, MGD follows the general distillation paradigm.
Given batches of data fed into teacher and student, MGD matches their intermediate channels from distillation position.
The motivation is that whether the student has been pre-trained or not, each channel of it should be guided by its high related teacher channel directly to narrow the semantic feature gap.
In order to implement element-wise losses, teacher channels would be reduced according to the matching graph.
The method for channel reduction is flexible, we propose three manners: sparse matching, random drop and absolute max pooling.
Experiments show that all three ways are effective on various tasks.
For the whole training, we apply coordinate descent algorithm~\cite{wright2015coordinate} to alternate between two optimizations for channels matching and weights updating.
Furthermore, distilling a pre-trained student using MGD is efficient due to its parameter-free nature as same as training from-scratch.
\section{Related Work}
\noindent\textbf{Correspondence Problem.}
Finding optimal correspondence between two sets of instances is a crucial step for a wide range of computer vision problems, such as shape retrieval~\cite{huet1999graph}, image retrieval~\cite{rubner2000earth}, object categorization~\cite{duchenne2011graph} and video action recognition~\cite{brendel2011learning}.
Linear Assignment (LA) is the most classical correspondence problem that can be efficiently solved with Hungarian algorithm~\cite{kuhn1955hungarian}, unlike the NP-hard quadratic assignment problem~\cite{zhou2012factorized}.
Matching based training losses~\cite{frogner2015learning,huang2017like} contain the ideology of matching features, but they have a heavy computation.
For example, the Wasserstein loss~\cite{frogner2015learning} uses the iterated optimization~\cite{cuturi2013sinkhorn} to approximate the matching matrix, its computation cost will dramatically become large along with the growth of feature dimensions.
It only can be used for the feature from the last fully-connected layer, so does~\cite{huang2017like}.
In this paper, we treat relationships modeling between teacher and student channels as a LA problem, particularly the min cost assignment problem.
The total matching cost function would be minimized by the Hungarian method to achieve a bipartite assignment graph.
This graph represents the high related channel pairs between teacher and student feature sets.

\noindent\textbf{Knowledge Distillation.}
Pioneered by \cite{hinton2015distilling}, the classic method for knowledge distillation contains two constituents: logits from the last teacher layer used as the soft targets, and Kullback-Leibler divergence loss used to let student match these targets.
However, the performance of output distillation is limited due to the very similar supervised signal from teacher model with ground-truth.
More works~\cite{romero2014fitnets,yim2017gift,chen2017learning} switch to feature distillation by combining intermediate features together to strongly supervise the student.
All these works rely on certain adaptation modules between hidden layers, in order to solve the contrast of semantic feature structures.
Feature correlation based methods provide more fine-grained recipes to perform knowledge distillation, such as attention transfer~\cite{Zagoruyko2017AT,lee2019graph}, neuron selectivity transfer~\cite{huang2017like}.
These works focus on capturing and transferring the spatial information for intermediate feature maps.
Another technique fashion for distillation is to design the loss function, including activation transfer loss with boundaries formed by hidden neurons~\cite{heo2019knowledge}, the loss for penalizing structural differences in relations~\cite{wonpyo2019rkd}.
In this paper, we propose a novel perception that performing distillation after matching intermediate channels between teacher and student.
Our proposed approach is intuitive and lightweight. It introduces marginal computation costs during training.
In the end, although the work of Jacobian matching-based distillation~\cite{srinivas2018knowledge} seems related to ours, it still suffers from the problems discussed in Section.~\ref{sec:intro}.
Because it not only uses adaption modules but also a specialized loss function.
 
\noindent\textbf{Transfer Learning.}
Commonly fine-tuning is one of the effective methods for knowledge transfer learning.
The student has been already pre-trained on a specific domain data, and then it's fine-tuned for another task with priori knowledge.
The work~\cite{he2019rethinking} finds that the model with random initialization could be trained no worse than using pre-trained parameters.
However, the model may suffer from the low capacity trained on a small dataset like Caltech-UCSD Birds 200~\cite{WelinderEtal2010}.
A number of previous works use ImageNet pre-trained models on different tasks such as detection and segmentation~\cite{wu2019detectron2}.
In this paper, we show that MGD can be also used for knowledge transfer learning with a pre-trained student for the further performance improvements in another task, particularly for the fine-grained categorization.
\section{Methodology}
\begin{figure*}[t]
  \centering
  \includegraphics[width=1.0\linewidth]{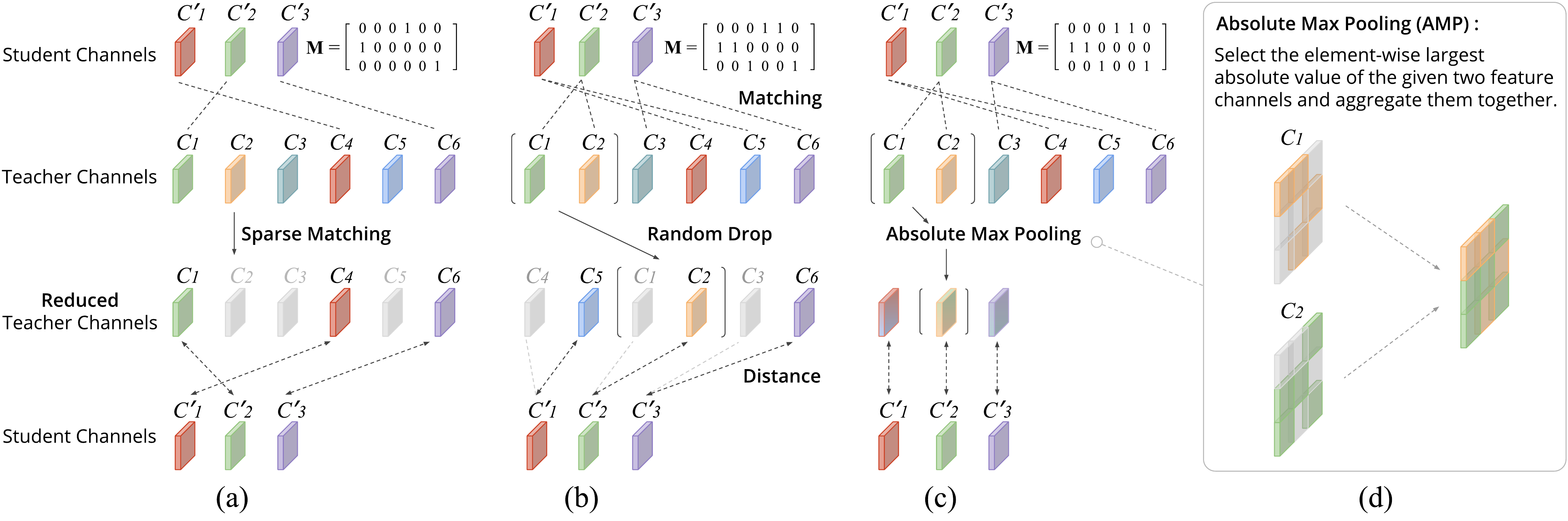}
  \caption{\small{
  \textbf{Channel reduction methods}.
    ${C_i}$ indicates teacher channels of intermediate features as same as $C'_{i}$ for student. 
    $\M$ is the matching matrix.
    We propose three effective methods to play reduction for teacher channels:
    sparse matching, random drop and absolute max pooling.
    (a) Sparse Matching.
    Each student channel only matches one teacher channel.
    Unmatched teacher channels are directly ignored.
    (b) Random Drop.
    Each student channel matches more than one teacher channel.
    Matched teacher channels with the same student channel are \textit{randomly} dropped to leave just one for guiding student.
    (c) Absolute Max Pooling (AMP).
    Each student channel matches more than one teacher channel.
    AMP picks out the element-wise largest absolute value along the channel axis over a group of matched teacher channels with the same student channel.
    (d) shows the detail about how AMP works on two channel tensors.
  }}
\label{fig:channel reduction}
\end{figure*}
In this section, we introduce a general formulation of the proposed Matching Guided Distillation (MGD).
MGD consists in a parameter-free channel matching module that can guide to shave teacher channels in three effective manners: sparse matching, random drop and absolute max pooling, as shown in Fig.~\ref{fig:channel reduction}.
\subsection{Feature Distillation Revisit}
We begin by briefly reviewing feature distillation in general formulation.
Suppose that 2D images\footnote{Bold capital letters denote a matrix
  $\X$, bold lower-case letters a column vector $\x$. $\x_i$ and $\x^j$
  represents the $i^{th}$ column and $j^{th}$ row of the matrix $\X$ respectively. $x_{ij}$ or
  $[\X]_{ij}$ denotes the scalar in the $i^{th}$ row and $j^{th}$
  column of the matrix $\X$. All non-bold letters represent
  scalars.} $\X$ are fed into the teacher $f_T$ and student $f_S$ networks that generate intermediate feature sets
\begin{aligns}
\label{eq:extract features}
  \T &= f_T(\X) \in \real^{C_T \times N}, \ \bS = f_S(\X) \in \real^{C_S \times N},
\end{aligns}
respectively at the target distillation positions.
Without loss of generality, we assume the feature maps are of the same spatial size $N=HW$ (height $H$ and width $W$) but could consist of different number of channels, e.g. $C_T=512$ while $C_S=128$.
Given a teacher network $f_T$ with frozen parameters, we wish to enhance the training of the student networks $f_S$ using the hints from $f_T$.
In a nutshell, the problem of feature distillation seeks for the optimum student network $f_S$ that minimizes the loss of the main task together with feature discrepancy penalty:
\begin{aligns}
   Loss = L_{task} + \gamma L_{distill},
\end{aligns}
where $\gamma$ is a trade-off coefficient for distillation loss.
The task loss $L_{task}$, for example, could be cross-entropy loss for classification or smooth-$L_1$ loss for object localization.
The key to feature distillation is the design of the distillation loss, $L_{distill}$, which ensures the similarity between intermediate features $\T$ and $\bS$:
\begin{aligns} 
\label{eq:distill}
  L_{distill} = d_p(\sigma_T(\rho(\T, \M)), \sigma_S(\bS)),
\end{aligns}
where $\sigma_T,\sigma_S,d_p$ are the teacher and student feature transforms that convert raw feature into an easy-to-transfer form and distance functions respectively.
In the past few years, various designs have been proposed to make better use of information contained in teacher networks.
In MGD, $\sigma_T$ is a marginal ReLU as same as~\cite{heo2019overhaul}.
Based on a recent comprehensive review~\cite{heo2019overhaul} on these design aspects, we build the pipeline by employing the marginal ReLU for teacher transform $\sigma_T$:
\begin{aligns}
  \sigma_T(x) = \max(x, m),
\end{aligns}
where $m < 0$ is a margin value, computed as an expectation value over all training images.
Following~\cite{heo2019overhaul}, we choose the partial $L_2$ loss function to calculate feature distance:
\begin{aligns}
  d_{p}(\T, \bS) = \sum_{i}^{C} \sum_{j}^N
  \bigg\{
  \begin{array}{ll}
    0 & \text{if} \ \ s_{ij} \leq t_{ij} \leq 0
    \\
    (t_{ij} - s_{ij})^2 & \text{otherwise},
  \end{array}
\end{aligns}
for any pair of matrices $\T, \bS \in \real^{C \times N}$ of the same dimension.
\subsection{Channel Matching}
The distillation loss (Eq.~\ref{eq:distill}) plays a vital role in distilling the knowledge of a complex model into a simpler one.
To achieve this goal, the design of distance function ($d_p$) and feature transforms ($\sigma_T, \sigma_S$) needs to ensure teacher's knowledge can be transferred to student with minimum loss.
Despite that various choices have been proposed in the past few years (see \cite{heo2019overhaul} for an extensive review), it is still necessary to add an $1 \times 1$ convolutional layer or other module on student ($\sigma_S$) to bridge the semantic gap between $\T$ and $\bS$.
The presence of student transform not only adds burden on network complexity but also complicates the training procedure.
We propose MGD by completely removing the student transform from the distillation pipeline, i.e. $\sigma_S(\bS) = \bS$.
Instead, we directly match the channel via the reduction operation $\rho(\T, \M)$ from teacher to student.
This operation is parameter-free and efficient to optimize in training.
Below we explain how to establish the correspondence $\M$ across channels and define the implementation of $\rho(\cdot, \cdot)$ in next section.

Given a pair of teacher feature $\T \in \real^{C_T \times N}$ and student one $\bS \in \real^{C_S \times N}$, we first encode their pairwise relation in a distance matrix, $\D \in \real^{C_S \times C_T}$, whose element $d_{ij}$ computes the Euclidean distance between $i^{th}$ student and $j^{th}$ teacher channels:
\begin{aligns}
  d_{ij} = \sum_{k=1}^{N} (s_{ik} - t_{jk})^2.
\end{aligns}
Our goal is to find a binary matrix $\M \in \{0,1\}^{C_S \times C_T}$ that encodes the channel-wise correspondence, where $m_{ij} = 1$ if $i$-th student and $j$-th teacher channels are pertinent.
In the special case when the teacher and student feature maps are of the same dimension (i.e., $C_T = C_S$), the matching is assumed to be one-to-one and the resulting matrix $\M'$ defines a permutation of $C_T$ channels:
\begin{aligns}
\label{eq:constraint2}
\Pi = \Big\{\M' \in \bina^{C_T \times C_T} | \sum_{j=1}^{C_T}{m'_{ij}} = 1, \sum_{i=1}^{C_T}{m'_{ij}} = 1 \Big\}.
\end{aligns}
In general, we resort to a many-to-one matching as the teacher channel number $C_T$ is often several times more than the student one $C_S$.
In order to make the distillation procedure evenly distributed over feature channels, we further constrain that each student channel has to be associated with $\alpha=\lfloor C_T/C_S \rfloor$ teacher channels.
More specifically, the many-to-one balanced matching $\M$ satisfies:
\begin{aligns}
\label{eq:constraint}
\Pi_b = \Big\{\M \in \bina^{C_S \times C_T} | \sum_{i=1}^{C_S}{m_{ij}} = 1, \sum_{j=1}^{C_T}{m_{ij}} = \alpha \Big\}.
\end{aligns}
This constraint enforces that $\M$ is a wide-shape matrix, where the sum of each column equals to one because each teacher channel can only be connected to one student channel.
On the other hand, the sum of each row needs to be $\alpha$.
In another word, each student channel has to be associated with $\alpha$ teacher ones.

Given two sets of feature channels with the associated pairwise distance, the problem of channel matching consists in finding a balanced many-to-one mapping $\M$ such that the sum of matching cost is minimized:
\begin{aligns}
\begin{split}
   \label{eq:lp}
   \min_{\M} \text{trace}(\D^T \M) = \sum_{i=1}^{C_S} \sum_{j=1}^{C_T} d_{ij} m_{ij}, \ \text{subject to} \ \M \in \Pi_b.
\end{split}
\end{aligns}
Now Eq.~\ref{eq:lp} is not a standard linear assignment problem, because the Hungarian algorithm works on the square cost matrix.
To satisfy this prerequisite, let $\D' = [\D; \cdots; \D] \in \real^{C_T \times C_T}$ to be a square matrix by concatenating $\alpha$ matrices $\D$ vertically\footnote{
  For the case when $C_T$ is not divisible by $C_S$, we simply shave $C_T - \alpha C_S$ dummy teacher channels off the cost matrix. Although this solution is not optimal, we found the result is still promising on several datasets.
}.
Our solution proceeds by optimizing a standard linear assignment problem to solve out $\M'$ first:
\begin{aligns}
\begin{split}
   \label{eq:lp2}
   \min_{\M'} \text{trace}(\D'^T \M') = \sum_{i=1}^{C_T} \sum_{j=1}^{C_T} d'_{ij} m'_{ij}, \ \text{subject to} \ \M' \in \Pi,
 \end{split}
\end{aligns}
using Hungarian algorithm.
We then evenly slice the resulting matrix $\M' = [\M_1'; \cdots; \M'_{\alpha}]$ in row blocks, where each sub-matrix $\M'_i \in \bina^{C_S \times C_T}$ is of the same size.
It's easy to prove that the optimal solution for Eq.~\ref{eq:lp} is:
\begin{aligns}
\label{eq:sol}
   \M = \sum_{i=1}^\alpha \M'_i \in \bina^{C_S \times C_T}.
\end{aligns}
\subsection{Channel Reduction}
Once the channel-wise correspondence $\M$ is established, the distillation loss $d_p$ in Eq.~\ref{eq:distill} would encourage the student channel to mimic the hidden feature of the related teacher channels.
Because of the many-to-one nature for the mapping, we discuss below three parameter-free choices for reducing teacher feature via operation $\rho(\T, \M)$ to match with student feature.
\\[5pt]
\noindent\textbf{Sparse Matching.}
To match the $C_T$ teacher channels with $C_S$ student ones, the straightforward way is to pick an optimal subset of $C_S$ teacher channels and construct a one-to-one matching with the $C_S$ student channels.
To do so, we simply formalize another linear assignment problem by introducing $C_T - C_S$ dummy student channels, each of which is put in an infinity distance from any teacher channel.
This linear assignment problem is in the same form as Eq.~\ref{eq:lp2} except the distance matrix $\D' \in \real^{C_T \times C_T}$ is constructed by appending $C_T - C_S$ rows of large constant (e.g. $1e10$) to the end of the original $\D \in \real^{C_S \times C_T}$.
After applying the Hungarian algorithm on Eq.~\ref{eq:lp2}, we could find for each student channel the most relevant teacher one, which are encoded in the first $C_S$ rows of the resulting correspondence matrix, i.e., $\M = \M'_1 \in \bina^{C_S \times C_T}$.
For instance, Fig.~\ref{fig:channel reduction}(a) illustrates an example of matching $C_T=6$ teacher channels with $C_S=3$ student ones, where the correspondence matrix $\M \in \bina^{3 \times 6}$ denotes three one-to-one matching pairs.
In this case, the reduction operation can thus be defined as:
\begin{aligns}
\label{eq:sm}
  \rho_{\text{SM}}(\T, \M) = \M \T \in \real^{C_S \times N}.
\end{aligns}
\noindent\textbf{Random Drop.}
The major limitation of the first sparse matching choice is that it only retains a small fraction ($C_S/C_T$) of information conveyed in the original teacher features.
To reduce the information loss, our second choice for teacher reduction is to sample a random teacher channel from the ones associated with each student channel.
More specifically, there are $\alpha$ non-zero elements in each row of $\M$ according to the constraint (Eq.~\ref{eq:constraint}).
The random drop operation modifies the correspondence as $\M^{RD} \in \bina^{C_S \times C_T}$ by randomly keeping one non-zero element in each row, i.e., $\sum_{j=1}^{C_T} m^{RD}_{ij}=1$ for any $i=1,\cdots,C_S$.
To have a better understanding, we visualize one case in Fig.~\ref{fig:channel reduction}(b), where the second student channel $C'_2$ is associated with $C_1$ and $C_2$ teacher channels after the channel matching step.
In random drop reduction, we randomly pick one of them (e.g. $C_2$) to match with the student.
In order to maximize the randomness, we generate correspondence matrices $\M_i^{RD}$ independently for different spatial positions of the feature map.
The overall reduction operation can be defined as:
\begin{aligns}
\label{eq:rd}
  \rho_{\text{RD}}(\T, \M) = \Big[ \M^{RD}_{1} \bt_1, \cdots, \M^{RD}_{N} \bt_N \Big] \in \real^{C_S \times N},
\end{aligns}
where $\bt_j \in \real^{C_T}$ denotes the $j^{th}$ column of the feature $\T$.
\\[5pt]
\noindent\textbf{Absolute Max Pooling.}
Following~\cite{heo2019overhaul}, we place the distillation module before ReLU.
Therefore both positive and negative values are transferred from teacher via the partial distance loss $d_p$ to student.
We hope the reduced teacher features still take the maximum activations in the same spatial position, including positive (usable) and negative (adverse) information.
To reach this purpose, we propose a novel pooling mechanism, named Absolute Max Pooling (AMP), as shown in Fig.~\ref{fig:channel reduction}(c).
Given a set of feature activations $\x = [x_1, \dots, x_C]^T \in \real^{C}$, the AMP is designed to choose the element that yields the largest magnitude:
\begin{aligns}
\label{eq:amp}
   f_{\text{AMP}}(\x) = \argmax_{x_i} | x_i |.
 \end{aligns}
 Similar to the random drop idea, the AMP operation is performed independently for each spatial position $\bt_j \in \real^{C_T}$ of the teacher feature map $\T \in \real^{C_T \times N}$.
 For each student channel $i = 1, \cdots, C_S$, AMP is used to select the most active teacher channel among the associated $\alpha$ ones.
 Because this teacher-student association has been encoded as the non-zero elements of $i^{th}$ row $\m^i \in \bina^{\C_T}$ of matrix $\M$, we can write the overall reduction operation in matrix form as:
\begin{aligns}
\label{eq:amp reduction}
  \rho_{\text{AMP}}(\T, \M) = \Big[ f_{\text{AMP}} (\m^{i} \circ \bt_j) \Big]_{ij} \in \real^{C_S \times N},
\end{aligns}
where $\circ$ indicates the element-wise product between two vectors.
As shown in Fig.~\ref{fig:channel reduction}(d), AMP pools these $\alpha$ feature nodes into single one along the channel axis.
\\[5pt]
\noindent\textbf{Theoretical Summary.}
In the theoretical perspective, we describe the understanding on these three channel reduction methods as following. 
All these methods belong to variants of the LA problem. 
With the same matrix notation, their goal can be connected as seeking for the optimal matching matrix $\M$ by optimizing certain linear objective. The first sparse matching (Eq.~\ref{eq:sm}) is a simple modification of the original Hungarian algorithm. 
The second random drop (Eq.~\ref{eq:rd}) and the last absolute max pooling (Eq.~\ref{eq:amp}) ideas progressively improve sparse matching by different approaches to reduce information loss from $\sigma_\T$.
To better understand and verify this insight, we have done many experiments and ablation studies in Section.~\ref{sec:experiments}.
\subsection{Implementation Details}
\begin{figure}[h]
\begin{minipage}[]{0.5\textwidth}
  \centering
  \includegraphics[width=1.\textwidth]{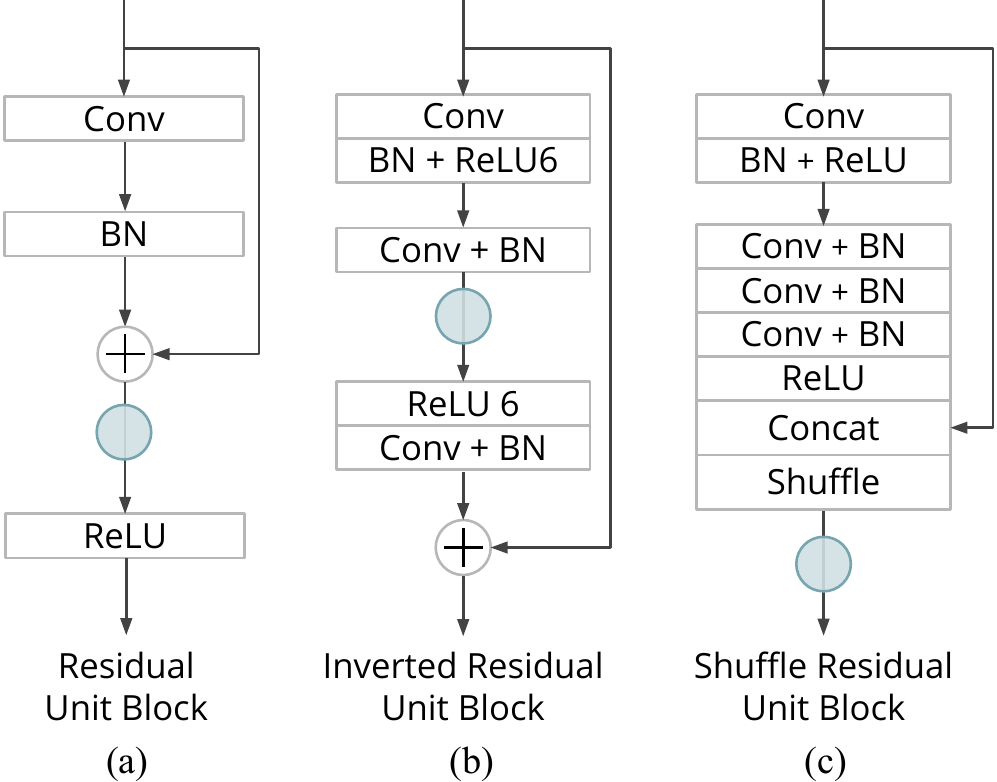}
\end{minipage}
\hskip 4pt plus 0fil
\begin{minipage}[]{0.477\textwidth}
  \vskip -17pt plus 0fil
  \centering
  \caption{\small{
    The distillation position in a unit block is pinpointed by a {\color{cyan}{color}} point.
    (a) For the standard residual unit in ResNet and MobileNet-V1, we use the output before ReLU.
    (b) For the inverted unit in MobileNet-V2, we use the intermediate feature before ReLU6.
    Although the output from the whole block isn't activated by ReLU-like operations too, the channel number is narrowed down into an unusable one for distillation.
    (c) The feature after channel shuffle is used for the shuffle unit.
  }}
  \label{fig:distillation position}
\end{minipage}
\vskip -10pt plus 0 fil
\end{figure}

\noindent\textbf{Distillation Position.}
In our experiments, we use contemporary models in recent years, including ResNet~\cite{he2016deep}, MobileNet-V1~\cite{howard2017mobilenets}, -V2~\cite{sandler2018mobilenetv2} and ShuffleNet-V2~\cite{ma2018shufflenet}.
Commonly, these models contains four stages, each of which is composed of repeated unit blocks, as shown in Fig.~\ref{fig:distillation position}.
We apply distillation in the last unit block of each stage.
In the cases of distilling MobileNet-V2 and ShuffleNet-V2, the first stage is skipped, because their first stage is only a convolutional layer and also its feature map size ($H \times W$) is distinct from teacher's.
%
%
\\[5pt]
\noindent\textbf{Coordinate Descent.}
To optimize the whole system (Eq.~\ref{eq:distill}), we use Coordinate Descent algorithm~\cite{wright2015coordinate} by alternating between solving the combinatorial matching problem and updating network weights.
Postulating the matching is solved, we plug $\M$ in Eq.~\ref{eq:distill} and employ Stochastic Gradient Descent (SGD) to update student network weights $f_S$.
After several epochs or iterations of training with SGD, the student is switched into evaluation mode without learning.
Then we feed a dataset that is randomly sampled from \textit{training} data into student and teacher, in order to update matching $\M$ that optimizes Eq.~\ref{eq:lp} for the next training rounds.
Solving Eq.~\ref{eq:lp} takes the computational complexity of $O(C_T^3)$. 
This step introduces negligible cost in our implementation.
\section{Experiments}
\label{sec:experiments}
\textbf{Datasets.} 
We run a number of major experiments to compare our MGD with other methods, using sparse matching (SM), random drop (RD) and absolute max pooling (AMP).
We evaluate them in multiple tasks, including large-scale classification, fine-grained recognition with transfer learning, detection and segmentation.
We have done on four popular open datasets.
\textbf{CIFAR-100}~\cite{krizhevsky2009learning} is composed of 50,000 images within 100 classes, and has a fixed input size of $32 \times 32$.
ImageNet (\textbf{IN-1K})~\cite{deng2009imagenet} has 1.2 million training images and 50,000 validaton images in 1000 object categories.
Birds-200-2011 (\textbf{CUB-200})~\cite{WelinderEtal2010} is a dataset for categorizing the fine-grained objects, which contains 11,788 images of 200 bird classes.
CUB-200 is used for testing MGD in transfer learning.
\textbf{COCO}~\cite{lin2014microsoft} is a standard object detection and instance segmentation benchmark.
\\[5pt]
\noindent\textbf{Experimental Setting.}
Classification tasks use a standard training scheme.
For the pre-trained student, we set init learning rate (LR) in 0.01, otherwise 0.1 for training from scratch.
On CIFAR-100, the number of total epochs is 200, LR is dropped twice by 0.1 at $100^{\text{th}}$ and $150^{\text{th}}$ epoch.
On IN-1K and CUB-200, total epochs number is 120, and LR is dropped by 0.1 after every 30 epochs.
Momentum is 0.9 and weight decay is 5e-4.
We randomly crop 224 from 256 and then perform horizontal flip in IN-1K and CUB-200.

For object detection and instance segmentation, we follow the configurations in {\fontfamily{qcr}\selectfont Detectron2}~\cite{wu2019detectron2}.
The input size of training image is restricted in maximum size of 1333 and minimal size of 800.
Horizontal flip is the only data augmentation.
We use 8 GPUs and set mini-batch size of 2 images for each GPU.
The total iterations number is 90k.
We use standard 1x and 2x training schedules settings.

The number of images for solving $\M$ depends on the dataset scale.
We randomly sample 20k images on IN-1K and use all the training images on CIFAR-100 and CUB-200.
In the detection and segmentation task, about 5k images are used for updating $\M$.
\subsection{Main Results}
\noindent\textbf{Classification.}
Following the competitor~\cite{heo2019overhaul}, student models are randomly initialized in tasks on CIFAR-100.
We use WideResNet (WRN)~\cite{zagoruyko2016wide} as the teacher, which is in the model setting of depth 28 and wide factor 4, indicated by s:28-4.
The student has multiple settings in different compression aspects: depth, wide factor, and architecture.
Results comparison is shown in Table~\ref{table:results on cifar-100}.
WRN with s:16-4 has the same wide factor but smaller depth, so its channels are in same number of teacher's at the same distillation positions.
Since there is no need to reduce teacher channels, we only use MGD-SM to achieve the lowest error rate $20.53\%$.
WRN with s:28-2 has the same depth but a larger wide factor, thus teacher channels need to be reduced.
The results show that MGD-SM is better than~\cite{heo2019overhaul} with $+0.76\%$, MGD-RD is a little worse with $+0.23\%$, and MGD-AMP achieves the lowest error of $21.12\%$.
In experiments with s:16-2, whose both depth and wide factor are smaller than those of teacher, MGD-SM \& RD are worse than the competitor and MGD-AMP is competitive.
Another setting for student is using a different architecture, ResNet with s:56-2, which is deeper but with the same wide factor.
MGD has the competitive results under this setting as well.
The Table~\ref{table:results on cifar-100} shows that MGD can handle various compression types for student.
%
%
\begin{table}[h]
  \centering
  \scriptsize
  \begin{tabular}[h]{llc|lc|lc|lc}
    \cmidrule{1-9}
    model setting & method & error. & model setting & error. & model setting & error. & model setting & error. \\
    \toprule
    WRN   & Teacher & 21.09 & - & 21.09 & - & 21.09 & - & 21.09 \\
    w/ s:28-4	& & & & & & & & \\
    \cmidrule{1-9}
    WRN   & Baseline & 22.72 & WRN & 24.88 & WRN & 27.32 & ResNet & 27.68 \\
    w/ s:16-4   & KD~\cite{hinton2015distilling}  & 21.69 & w/ s:28-2 & 23.43 & w/ s:16-2 & 26.47 & w/ s:56-2 & 26.76 \\
    & FitNets~\cite{romero2014fitnets}            & 21.85 && 23.94 && 26.30 && 26.35 \\
    & AT~\cite{Zagoruyko2017AT}                   & 22.07 && 23.80 && 26.56 && 26.66 \\
    & Jacobian~\cite{srinivas2018knowledge}       & 22.18 && 23.70 && 26.71 && 26.60 \\
    & AB~\cite{heo2019knowledge}                  & 21.36 && 23.19 && 26.02 && 26.04 \\
    & Overhaul~\cite{heo2019overhaul}             & 20.89 && 21.98 && 24.08 && \textbf{24.44} \\
    & MGD - SM                                    & \textbf{20.53} && 21.22 && 24.72 && 25.20 \\
    & MGD - RD                                    & - && 22.21 && 24.64 && 26.01 \\
    & MGD - AMP                                   & - && \textbf{21.12} && \textbf{24.06} && 24.91 \\
    \cmidrule{1-9}
  \end{tabular}
  \vskip 2.1pt plus 1fil
  \caption{\small{
    \textbf{Comparison of error rates} (\%) with various model settings on CIFAR-100.
    We average 5 runs to report final results.
  }}
  \label{table:results on cifar-100}
  \vskip -15pt plus 0fil
\end{table}
%
%

In large-scale classification on IN-1K, as shown in Table~\ref{table:results on in-1k}, we use ResNet-152 to distill ResNet-50.
Since the channel number is identical in each stage of them, we only investigate MGD-SM.
The overall results from other works, except the Overhaul method, MGD beats other methods with maximum $1.45\%$ improvement in top-1 accuracy.
In the case of distilling MobileNet-V1 by ResNet-50, using MGD-SM has similar result with~\cite{heo2019overhaul}.
MGD w/ AMP is the best overall methods.
%
%
\begin{table}[h]
  %
  %
  \begin{minipage}[t]{0.5\linewidth}
  \centering
  \scriptsize
  \begin{tabular}[t]{llcc}
  \cmidrule{1-4}
  model           & method                            & top-1 err.      & top-5 err.   \\
  \toprule
  ResNet-152      & Teacher                           & 21.69           & 5.95   \\
  \cmidrule{1-2}
  ResNet-50       & Baseline                          & 23.85           & 7.13   \\
                  & KD~\cite{hinton2015distilling}    & 22.85           & 6.55   \\
                  & AT~\cite{Zagoruyko2017AT}         & 22.75           & 6.35   \\
                  & AB~\cite{heo2019knowledge}        & 23.47           & 6.94   \\
                  & Overhaul~\cite{heo2019overhaul}   & \textbf{21.65}  & 5.83   \\
                  & MGD - SM                          & 22.02           & \textbf{5.68}	\\
                  & & & \\
                  & & & \\
                  & & & \\
  \cmidrule{1-4}
  ResNet-50       & Teacher                           & 23.85           & 7.13   \\
  \cmidrule{1-2}
  MobileNet-V1    & Baseline                          & 31.13           & 11.24  \\
                  & KD~\cite{hinton2015distilling}    & 31.42           & 11.02  \\
                  & AT~\cite{Zagoruyko2017AT}         & 30.44           & 10.67  \\
                  & AB~\cite{heo2019knowledge}        & 31.11           & 11.29  \\
                  & Overhaul~\cite{heo2019overhaul}   & 28.75           & 9.66   \\
                  & MGD - SM         & 28.79           & 9.65   \\
                  & MGD - RD         & 29.55           & 10.02  \\
                  & MGD - AMP        & \textbf{28.53}  & \textbf{9.65} \\
  \cmidrule{1-4}
  \end{tabular}
  \vskip 2.1pt plus 1fil
  \caption{\small{
    \textbf{Comparison of error rates} (\%) with MGD and previous works in large-scale classification on IN-1K.
  }}
  \label{table:results on in-1k}
  \end{minipage}
  %
  %
  \begin{minipage}[t]{0.5\linewidth}
  \centering
  \scriptsize
  \begin{tabular}[t]{llcc}
  \cmidrule{1-4}
  model           & method                            & top-1 err.      & top-5 err.      \\
  \toprule
  ResNet-50       & Teacher                           & 20.02           & 6.06            \\
  \cmidrule{1-2}
  MobileNet-V2    & Baseline - FT                     & 24.61           & 7.56            \\
                  & Baseline - FS                     & 54.97           & 27.0            \\
                  & KD~\cite{hinton2015distilling}    & 23.52           & 6.44            \\
                  & AT~\cite{Zagoruyko2017AT}         & 23.14           & 6.97            \\
                  & AB~\cite{heo2019knowledge}        & 23.08           & 6.54            \\
                  & Overhaul~\cite{heo2019overhaul}   & 21.69           & 5.64            \\
                  & MGD - SM                          & 21.82           & 5.68            \\
                  & MGD - RD                          & 21.58           & 5.92            \\
                  & MGD - AMP                         & \textbf{20.64}  & \textbf{5.38}   \\
  \cmidrule{1-2}
  ShuffleNet-V2   & Baseline - FT                     & 31.39           & 10.9            \\
                  & Baseline - FS                     & 66.28           & 35.7            \\
                  & KD~\cite{hinton2015distilling}    & 28.31           & 9.67            \\
                  & AT~\cite{Zagoruyko2017AT}         & 28.58           & 9.29            \\
                  & AB~\cite{heo2019knowledge}        & 28.22           & 9.48            \\
                  & Overhaul~\cite{heo2019overhaul}   & 27.42           & 8.04            \\
                  & MGD - SM                          & 28.22           & 8.85            \\
                  & MGD - RD                          & 27.71           & 8.72            \\
                  & MGD - AMP                         & \textbf{25.95}  & \textbf{7.46}   \\
  \cmidrule{1-4}
  \end{tabular}
  \vskip 2.1pt plus 1fil
  \caption{\small{
    \textbf{Comparison of error rates} (\%) on CUB-200.
    The students are pre-trained on IN-1K.
    FT: fine-tune. 
    FS: from-scratch.
  }}
  \label{table:results on cub}
  \end{minipage}
  \vskip -20pt plus 0fil
\end{table}
%
%
\\[5pt]
\noindent\textbf{Transfer Learning.}
We use fine-grained categorization on CUB-200 to investigate distillation for transfer learning.
We implement MGD and our competitor Overhaul for using ResNet-50 to distill light students.
The teacher ResNet-50 has been pre-trained on IN-1K and then trained on CUB-200.
We use two prevailed lightweight models, MobileNet-V2 and ShuffleNet-V2.
Conspicuously, they have fewer parameters than ResNet-50.
The main results have been shown in Table~\ref{table:results on cub}.
We have two baselines for each student: one is trained from scratch (Baseline-FS), and the other is fine-tuned (Baseline-FT) from IN-1K.
We summarize the experimental results in two folds.

First, the results of two baselines show that transfer learning from a general data domain is helpful to the specific task.
Students pre-trained on IN-1K could bring $\sim$$30\%$ accuracy improvements at least.

Second, we adopt our three reduction methods of MGD to compare with Overhaul~\cite{heo2019overhaul}.
The experimental phenomenon of two students are same.
MGD-AMP beats the Overhaul with $1.05\%$ improvement and also makes MobileNet-V2 almost have the similar performance with teacher.
ShuffleNet-V2 seems difficult to be distilled, there is a unignored gap with teacher.
But MGD-AMP stably performs best to help ShuffleNet-V2 achieve the maximum top-1 error decrease from $31.39\%$ to $25.95\%$.

No matter which reduction method we use to accomplish distillation for transfer learning, all the experimental results show that MGD is more friendly for distilling a pre-trained student.
%
%
\begin{table}[t]
  %
  %
  \begin{minipage}[t]{0.49\linewidth}
  \scriptsize
  \begin{tabular}[t]{llcc}
  \cmidrule{1-4}
  backbone			& method                             & $\text{AP}^{\text{bbox}}$ & $\text{AP}^{\text{bbox}}_{\text{50}}$ \\
  \toprule
  ResNet-50  \ \ \ \ \ \ \	& Teacher                            & 36.37                 & 55.37 \\
  \cmidrule{1-2}
  ResNet-18      	& Baseline                           & 30.30                 & 47.53 \\
                  	& Overhaul~\cite{heo2019overhaul}    & 30.02                 & 46.95 \\
                   	& MGD - AMP                          & \textbf{31.15}        & \textbf{48.60} \\
  \cmidrule{1-2}
  MobileNet-V2      & Baseline                           & 26.54                 & 42.14 \\
                    & Overhaul~\cite{heo2019overhaul}    & 26.62                 & 42.01 \\
                    & MGD - AMP                          & \textbf{27.45}        & \textbf{43.10} \\
  \cmidrule{1-4}
  \multicolumn{4}{c}{(a) RetinaNet, \textbf{1}x schedule + \textbf{single}-scale}
  \end{tabular}
  \end{minipage}
  %
  %
  \begin{minipage}[t]{0.499\linewidth}
  \raggedleft
  \scriptsize
  \begin{tabular}[t]{llcc}
  \cmidrule{1-4}
  backbone             & method                             & $\text{AP}^{\text{mask}}$ & $\text{AP}^{\text{mask}}_{\text{50}}$  \\
  \toprule
  ResNet-50	 \ \ \ \ \ \ \	& Teacher						& 33.5                 & 54.1 \\
  \cmidrule{1-2}
  ResNet-18            & Baseline                           & 26.1                 & 43.7 \\
                       & Overhaul~\cite{heo2019overhaul}    & 26.3                 & 43.9 \\
                       & MGD - AMP                          & \textbf{26.9}        & \textbf{44.2} \\
  \cmidrule{1-2}
  MobileNet-V2         & Baseline                           & 27.1                 & 44.8 \\
                       & Overhaul~\cite{heo2019overhaul}    & 27.0                 & 44.8 \\
                       & MGD - AMP                          & \textbf{27.6}        & \textbf{45.1} \\
  \cmidrule{1-4}
  \multicolumn{4}{c}{(b) EmbedMask, \textbf{1}x schedule + \textbf{single}-scale}
  \end{tabular}
  \end{minipage}
  %
  %
  \begin{minipage}[t]{0.49\linewidth}
  \vskip 3pt plus 1fil
  \scriptsize
  \begin{tabular}[t]{llcc}
  \cmidrule{1-4}
  backbone					& method						& $\text{AP}^{\text{bbox}}$ & $\text{AP}^{\text{bbox}}_{\text{50}}$ \\
  \toprule
  ResNet-50  \ \ \ \ \ \ \	& Teacher						& 37.01                 & 56.03 \\
  \cmidrule{1-2}
  ResNet-18            & Baseline                           & 30.78                 & 47.88 \\
                       & Overhaul~\cite{heo2019overhaul}    & 30.26                 & 47.22 \\
                       & AT~\cite{Zagoruyko2017AT} 			& 30.54 				& 47.65 \\
                       & AB~\cite{heo2019knowledge}			& 31.32 				& 48.70 \\
                       & MGD - AMP                          & \textbf{31.38}        & \textbf{48.79} \\
  \cmidrule{1-4}
  \multicolumn{4}{c}{(c) RetinaNet, \textbf{1}x schedule + \textbf{multi}-scale}
  \end{tabular}
  \end{minipage}
  %
  %
  \begin{minipage}[t]{0.499\linewidth}
  \vskip 3pt plus 1fil
  \raggedleft
  \scriptsize
  \begin{tabular}[t]{llcc}
  \cmidrule{1-4}
  backbone				& method							& $\text{AP}^{\text{bbox}}$ & $\text{AP}^{\text{bbox}}_{\text{50}}$ \\
  \toprule
  ResNet-50  \ \ \ \ \ \ \	& Teacher						& 38.73                 & 56.72 \\
  \cmidrule{1-2}
  ResNet-18            & Baseline                           & 34.63                 & 53.08 \\
                       & Overhaul~\cite{heo2019overhaul}    & 34.42                 & 52.90 \\
                       & AT~\cite{Zagoruyko2017AT} 			& 34.43 				& 52.97 \\
                       & AB~\cite{heo2019knowledge}			& 34.92 				& 53.50 \\
                       & MGD - AMP                          & \textbf{35.10}        & \textbf{53.76} \\
  \cmidrule{1-4}
  \multicolumn{4}{c}{(d) RetinaNet, \textbf{2}x schedule + \textbf{multi}-scale}
  \end{tabular}
  \end{minipage}
  \vskip 2.1pt plus 1fil
  \begin{minipage}[t]{1.\linewidth}
  \centering
  \caption{\small{
    \textbf{Comparison of object detection and instance segmentation results}. 
    We distill lightweight backbones of RetinaNet for object detection (a, c, d) and EmbedMask for instance segmentation (b) on COCO.
    Here we experiment only with AMP because it's the best operation among three reduction manners.
  }}
  \label{table:detection results}
  \end{minipage}
  \vskip -20pt plus 0fil
\end{table}
%
%
\\[5pt]
\noindent\textbf{Object Detection \& Instance Segmentation.}
To verify the generalization of MGD, we extend it to object detection and instance segmentation.
RetinaNet~\cite{lin2017focal} is a modern one-stage detector, which has excellent performance in both precision and speed.
EmbedMask~\cite{YingEmbedMask} is a novel framework for instance segmentation, which utilizes embedding strategy to generate instance masks on a unified one-stage structure.
In this section, we experiment with RetinaNet and EmbedMask respectively, using three different backbones: ResNet-50 as teacher, ResNet-18 and MobileNet-V2 as students.
All these backbones are pre-trained on IN-1K.
We train the models on COCO {\fontfamily{qcr}\selectfont train2017} set and test them on {\fontfamily{qcr}\selectfont val2017}.
Baselines are trained without distillation.
As comparison, we also train with~\cite{heo2019overhaul,Zagoruyko2017AT,heo2019knowledge} under same configurations.
The main results are presented in Table~\ref{table:detection results}.
For object detection, in both cases of distilling ResNet-18 and MobileNet-V2, MGD has stable improvements by $0.47 - 0.91$ point.
In segmentation, in the case of ResNet-18, we freeze the first two backbone stages to avoid {\fontfamily{qcr}\selectfont OOM}.
MGD can bring about $0.8$ mAP point for ResNet-18 and $0.6$ point for MobileNet-v2, it outperforms the competitor.
These results prove that MGD works more stable than previous works in object detection and instance segmentation.
\subsection{Ablation Study}
\label{sec:ablation study}
\begin{figure*}[t]
  \begin{minipage}[t]{1.\textwidth}
  \centering
  \includegraphics[width=1.0\textwidth]{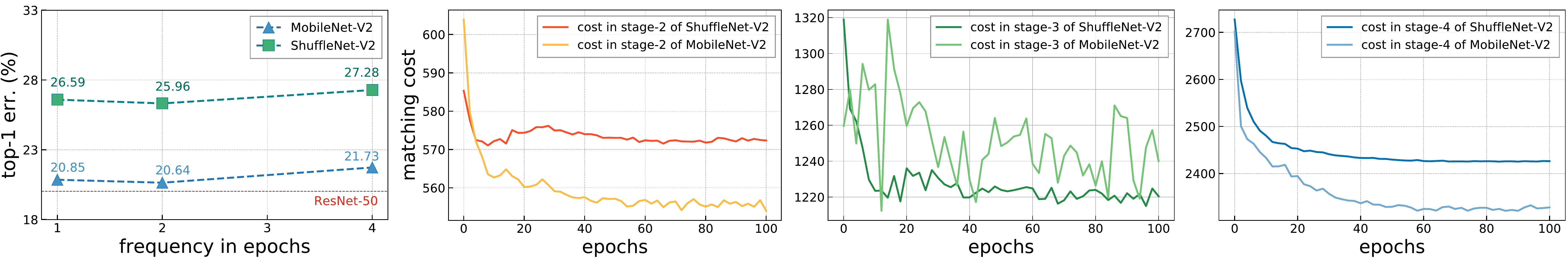}
  \end{minipage}
  \begin{minipage}[t]{.245\textwidth}
  \vskip -17pt plus 0fil
  \centering
  \caption{\small{
    Sensitivity to frequency of updating $\M$ on CUB-200.
  }}
  \label{fig:freq update}
  \end{minipage}
  \hskip 3pt plus 0fil
  \begin{minipage}[t]{.745\linewidth}
    \vskip -17pt plus 0fil
    \caption{\small{
      \textbf{Curves of matching cost} in three distillation stages of MobileNet-V2 and ShuffleNet-V2, which are distilled by ResNet-50 on CUB-200.
    }}
    \label{fig:cost curve}
  \end{minipage}
  \vskip -10pt plus 0fil
\end{figure*}

\noindent\textbf{Frequency of Updating $\M$.}
To find the best practices, we experiment on how intense the frequency of updating $\M$ could affect the MGD performance.
Here all experiments adopt MGD-AMP.
The baseline is updating $\M$ in the end of every training epoch (frequency=1) as same as validation does.
In Fig.~\ref{fig:freq update}, updating in every 2 training epochs achieves the best results both in MobileNet-V2 and ShuffleNet-V2 on CUB-200.
If the frequency becomes larger than 2 in epochs (frequency=4), it will produce higher error rates.
This study suggests that MGD should not be updated either fast or lazily.
On IN-1K, we update $\M$ after every epoch due to its large dataset scale.
%
%
%
%
\\[5pt]
\noindent\textbf{Absolute Max Pooling.}
Absolute max pooling (AMP) leads to largest improvements compared to proposed SM and RD.
Now we compare it with basic pooling operations, max pooling (MP) and average pooling (AvgP), to perform reduction along the channel axis.
Table~\ref{table:amp vs. mp} shows 
that AMP behaves stably better than MP and AvgP.
AvgP performs worst, because average pooling operation is easily to counteract the sharpness feature for a group of channel tensors. 
Albeit MP works closely with AMP, it's still not perfect because it would shave negative feature values by positive ones.
This result shows the effect of AMP for preventing feature information loss from reduction.
For a better illustration, we have a fundamental and intuitive experiment on these three pooling operations in 
Section.~\ref{sec:analysis of pooling operation}.
%
\\[5pt]
\begin{minipage}[t]{1.\linewidth}
	\begin{minipage}[t]{0.685\linewidth}
	\noindent\textbf{With \textit{vs.} Without Matching.}
	This ablation checks the importance of channel mathing mechanism.
	We remove matching process and simply use AMP as a feature reducer along channel axis.
	The right table shows 
	\end{minipage}
	\hfill
	\begin{minipage}[t]{0.29\linewidth}
	\vskip -5pt plus 0fil
	%
	%
		  \scriptsize
		  \begin{tabular}[t]{@{}l|c|c@{}}
		  \cmidrule{1-3}
		  model & \multicolumn{2}{c}{top-1 err.}              \\
		  \cmidrule{1-3}
		  with matching?  & 				& \checkmark      \\
		  \toprule
		  ResNet-50       & \multicolumn{2}{c}{20.02}         \\
		  \cmidrule{1-3}
		  MobileNet-V2    & 22.10           & \textbf{20.64}  \\
		  ShuffleNet-V2   & 27.62           & \textbf{25.95}  \\
		  \cmidrule{1-3}
		  \end{tabular}
	%
	\end{minipage}
	\begin{minipage}[t]{1.\linewidth}
	\vskip 3pt plus	0fil
	the results of MGD with and without channels matching.
	It proves the effectiveness of channels matching in MGD.
	Both of distilling MobileNet-V2 and ShuffleNet-V2 without matching are worse than that with matching about 1.5\% in top-1 error.
	\end{minipage}
\end{minipage}
\\[5pt]
\noindent\textbf{Capacity Analysis.}
Next, we illustrate MGD is more efficient for training than other methods.
We investigate the capacity of joint training with MGD and~\cite{heo2019overhaul}.
In experiments, we use four GeForce 1080Ti GPU cards to run training.
Under the same experimental settings, MobileNet-V2 has less parameters and memory consumption than teacher without distillation.
Table~\ref{table:capacity analysis} shows \cite{heo2019overhaul} brings too many parameters to cause {\fontfamily{qcr}\selectfont OOM}.
As well known, additional distillation module not only bring the learnable weights for training, but also additional gradients and optimizer states into GPU memory~\cite{pudipeddi2020training}.
In contrast, training with MGD has little bit of additional parameters due to its basic nature of parameter-free.
Moreover it has better results.
\\[5pt]
\noindent\textbf{Optimization Analysis.}
\begin{figure*}[t]
  \begin{minipage}[t]{1.\linewidth}
    \centering
    \includegraphics[width=1.0\linewidth]{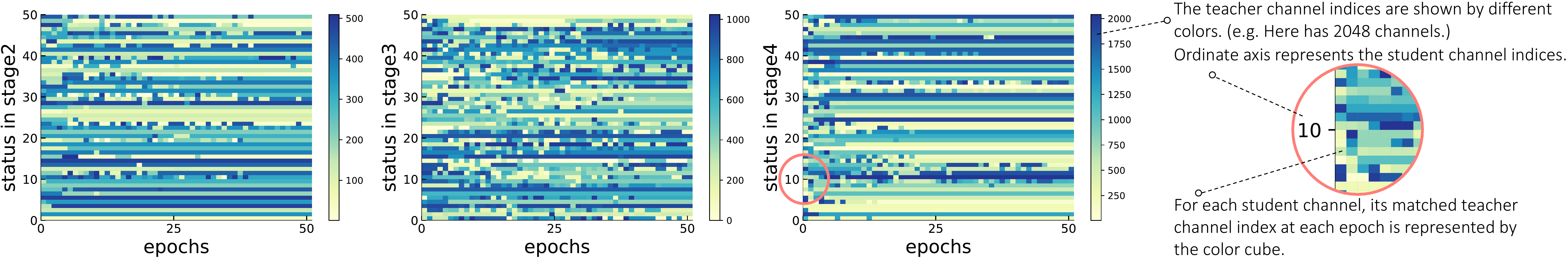}
  \end{minipage}
  \begin{minipage}[t]{1.\linewidth}
    \vskip -15pt plus 0fil
    \caption{\small{
      \textbf{Status of updating $\M$} in three distillation positions.
      Here we randomly select fifty student channels to check out their matched target channels, which are represented by the small cubes within different colors.
    }}
  \label{fig:matching matrices}
  \end{minipage}
\end{figure*}
%
%
%
%
\begin{table}[h]
  \begin{minipage}[t]{0.33\linewidth}
  \centering
  \scriptsize
  \begin{tabular}[t]{@{}l|c|c|c@{}}
  \cmidrule{1-4}
  model & \multicolumn{3}{c}{top-1 err.}            \\
  \cmidrule{2-4}
                & MP    & AvgP  & AMP               \\
  \toprule
  ResNet-50     & \multicolumn{3}{c}{20.02}         \\
  \cmidrule{1-4}
  MobileNet-V2  & 21.63 & 22.19 & \textbf{20.64}    \\
  ShuffleNet-V2 & 26.20 & 27.27 & \textbf{25.95}    \\
  \cmidrule{1-4}
  \end{tabular}
  \vskip 2.1pt plus 1fil
  \caption{\small{
    \textbf{Comparison of three pooling operations} for channel reduction.
  }}
  \label{table:amp vs. mp}
  \end{minipage}
  \hskip 5pt plus 0fil
  \begin{minipage}[t]{.65\linewidth}
  \centering
  \scriptsize
  \begin{tabular}[t]{lccccc}
  \cmidrule{1-6}
  model          & \textit{bs}  & method    & memory        & parameters      & top-1 err.      \\
  \toprule
  ResNet-50      & 256 & Teacher   & 8.10                  	& 25.1             & 20.02   \\
  \cmidrule{1-6}
  MobileNet-V2   & 256 & Student   & 5.01                  	& 3.5              & 24.61   \\
                 & 128 & Overhaul~\cite{heo2019overhaul}   	& {\fontfamily{qcr}\selectfont OOM}	& 31.5 & 21.69     \\
                 & 128 & MGD - AMP & 11.8                  	& \textbf{29.1}    & \textbf{20.64}   \\
  \cmidrule{1-6}
  \end{tabular}
  \vskip 2.1pt plus 1fil
  \caption{\small{
    \textbf{Capacity analysis.}
    Memory consumption is measured by gigabyte.
    Parameters is in millions.
    Here \textit{bs} indicates batch size.
  }}
  \label{table:capacity analysis}
  \end{minipage}
  \vskip -20pt plus 0fil
\end{table}
%
%
In this part, we analyze MGD in branches of visualization for understanding MGD with comprehensive vistas.
We set our experiments to check it in three aspects: matching cost, status of updating $\M$, and features matching \& reduction.

First, Fig.~\ref{fig:cost curve} shows the descent curves of matching cost in three distillation positions.
We track the sum of matching costs in every 2 epochs when distilling MobileNet-V2 and ShuffleNet-V2 with MGD-AMP.
The curves show that all the total costs are in the trend of descent during training.
This phenomenon is expected because the more related matched features are, the smaller their matching cost becomes.

Second, Fig.~\ref{fig:matching matrices} shows the updating status of $\M$ in distilling MobileNet-V2 on CUB-200.
Due to the massive channel number of intermediate features, we randomly select fifty student channels to visualize the updating status of $\M$.
All the three sub-figures have a common view that at the beginning, most of matching targets of each student channel change dramatically.
Then they will become stable after several training epochs.
This result concludes that coordinate descent is effective and friendly for the joint optimization with SGD.

Third, 
Fig.~\ref{fig:features matching} 
in Appendix checks out the intermediate results of MGD in multiple tasks.
In order to check the rightness of matching status, we use the intermediate features for visualization at the \textit{earlier} training iterations.
We can conclude the matching results can be trusted for guiding student to induce the better results.
\section{Discussion and Future Work}
We have presented MGD as an effective distillation method within the parameter-free nature, and evaluated its three channel reduction ways in various tasks.
We also experiment in multiple perspectives of ablation study to verify its effect.
In the future, it's possible to supervise student in a dense manner, for example, using more than four positions to perform distillation with MGD.
%


\clearpage
%
%
\bibliographystyle{splncs04}
\bibliography{egbib}
%
%
\clearpage
\renewcommand{\thefigure}{A\arabic{figure}}  
\renewcommand{\thesection}{A\arabic{section}}
\setcounter{section}{1}
\setcounter{figure}{0}
\setcounter{table}{0}
\section*{Appendix}
\begin{figure*}[ht]
  \includegraphics[width=1.0\linewidth]{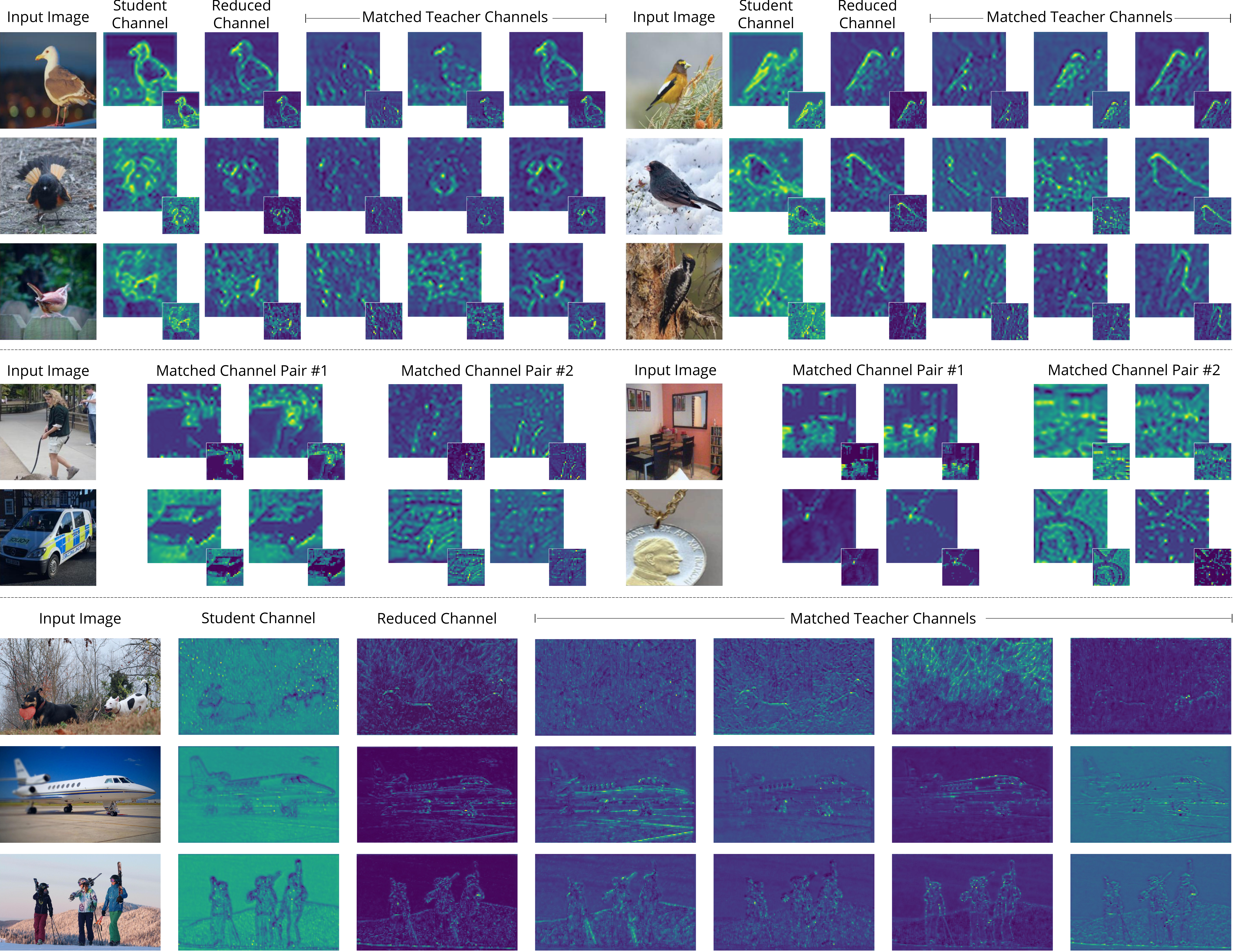}
  \caption{\small{
    \textbf{Channels matching with reduction.}
    The visualization has three parts separated by two dash lines.
    The first part (top) shows the matching results of \textit{stage-2} in MobileNet-V2 on CUB-200.
    The channel tensors are visualized in two square patches:
    small one is in original size of $28 \times 28$,
    the large one is generated by resizing small patch into the input size of $224 \times 224$.
    Each student channel matches three teacher channels.
    The second part (middle) shows the intermediate matching results in distilling ResNet-50 on IN-1K.
    Here we find the one-to-one match pair because student has the same channel number with teacher.
    We randomly select two pairs to visualize.
    The last part (bottom) shows the results in distilling ResNet-18 on COCO {\fontfamily{qcr}\selectfont train2017} set.
    Each student channel matches four teacher channels.
    According to this whole visualization, we can easily conclude that the semantic features activations are same between student channels and reduced channels generated by AMP operation.
  }}
  \label{fig:features matching}
  \vskip -15pt plus	0fil
\end{figure*}
%
%
\subsection{Analysis of Pooling Operations}
\label{sec:analysis of pooling operation}
\begin{figure}[t]
  \begin{minipage}[t]{1.\linewidth}
  \centering
  \includegraphics[width=1.\linewidth]{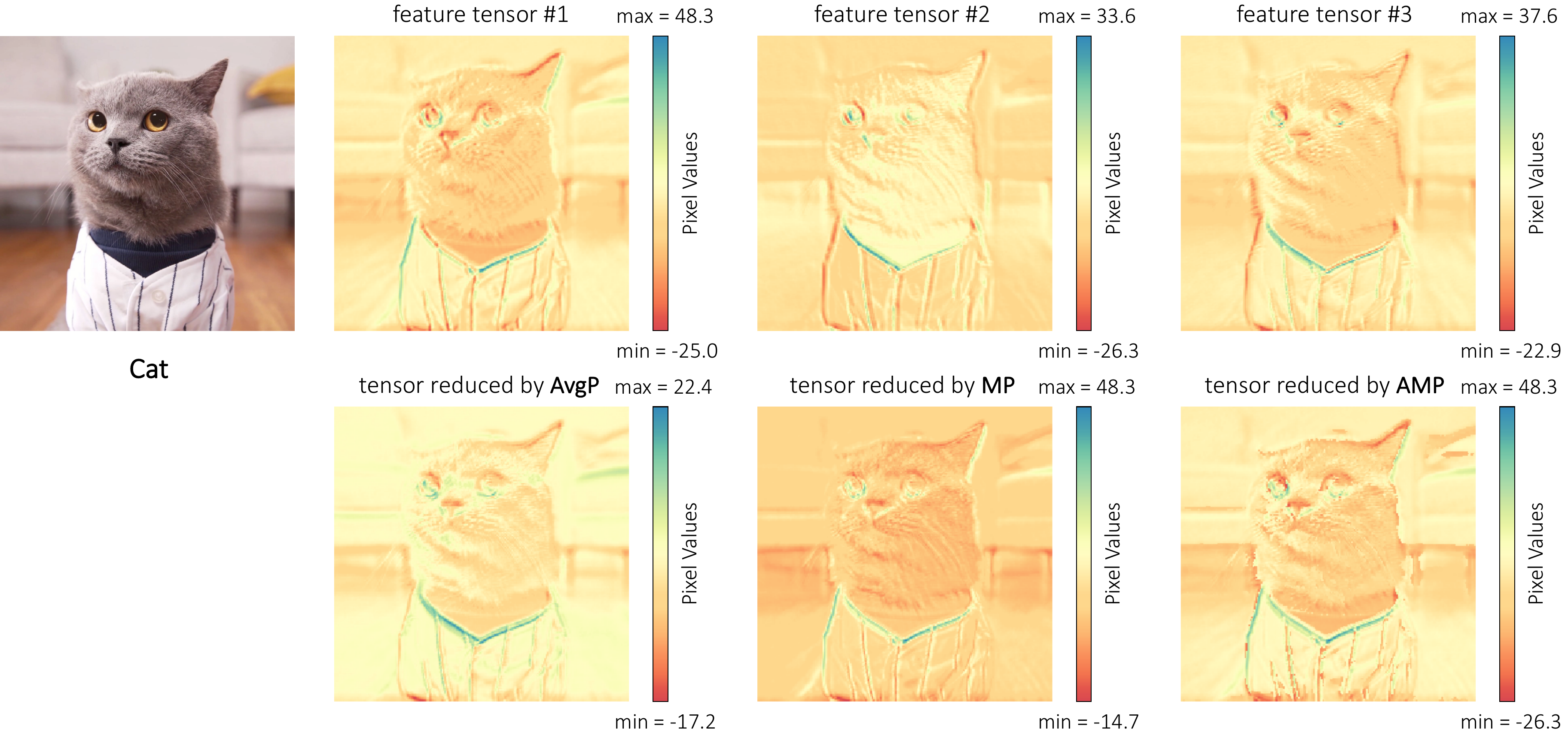}
  \end{minipage}
  \begin{minipage}[t]{1.\linewidth}
  \centering
  \includegraphics[width=1.\linewidth]{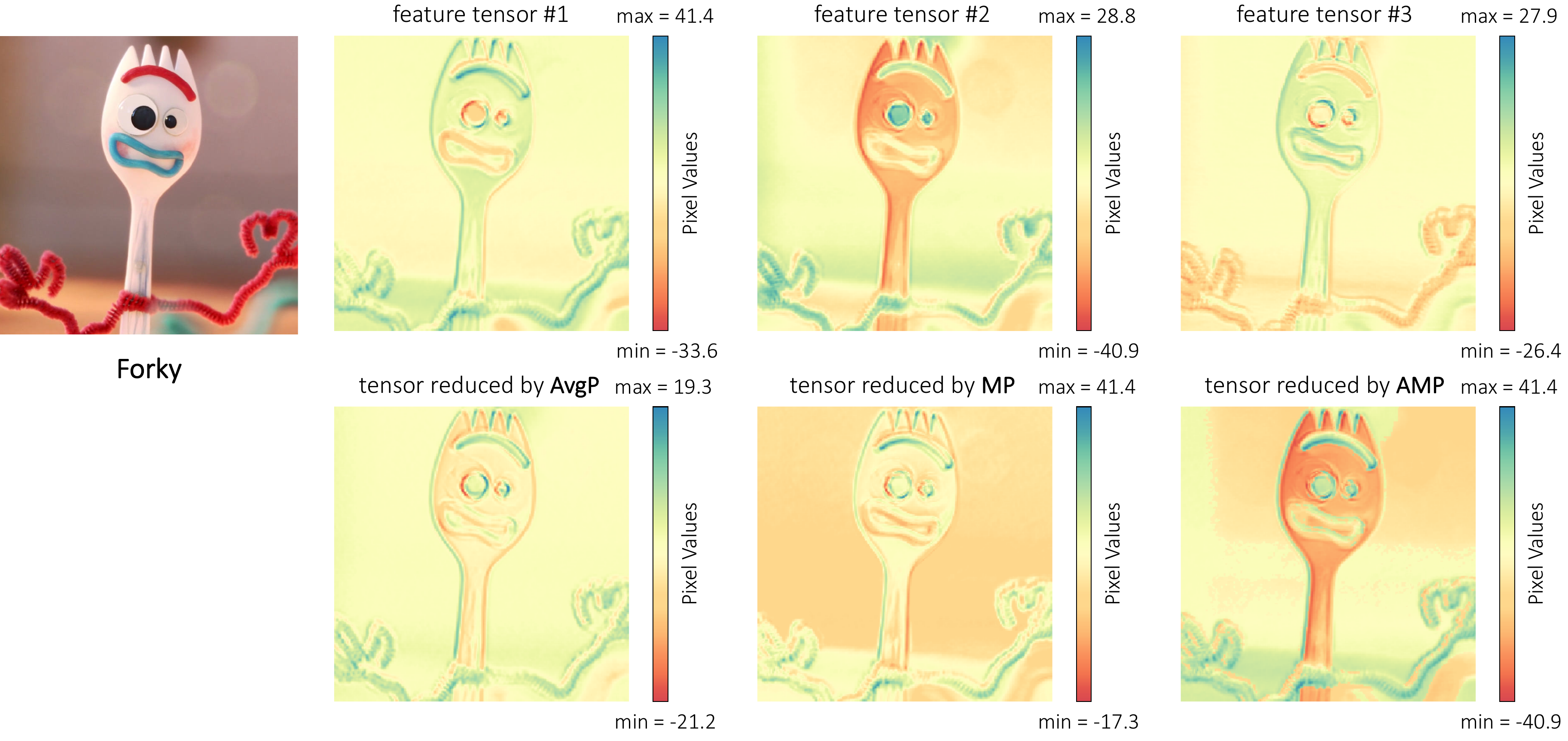}
  \end{minipage}
  \caption{\small{
    \textbf{Comparison of pooling operations}.
    All the feature tensors are normalized into $[0, 1]$ for visualization in order to clearly compare their textures in pixel level and degree.
    But their min and max values in the color bars use original pixel values without normalization.   
  }}
  \label{fig:pooling operations}
  \vskip -15pt plus	0fil
\end{figure}
In order to figure out why the absolute max pooling (AMP) stably works better than average pooling (AvgP) and max pooling (MP) when performing features reduction, we do a fundamental experiment in this part.
In Fig.~\ref{fig:pooling operations}, there are two input images (first column from left).
First, we build a very simple convolutional network (e.g. LeNet-7~\cite{lecun1998gradient}) with random initialization to extract features.
Then we select\footnote{
  This behavior imitates that three teacher channels have been matched with one student channel.
} three high-related tensors (in the same row with input images), which have similar semantic feature structures\footnote{
  The definition of similar feature structures is made according to their high responses in feature maps.
} with each other.
After using AvgP, MP and AMP operations to perform reduction, we achieve three reduced tensors of each example.

In the case of {\fontfamily{qcr}\selectfont Cat}, although AvgP keeps the responses of collar and eyes, it loses the edge activations of right shoulder.
MP works well, but its responses of eyes are too weak and also its responses of head texture (including background) are stronger than those of three original features.

In the case of {\fontfamily{qcr}\selectfont Forky}, AvgP erases the face-body responses from $2^{\text{th}}$ feature.
MP not only shades the negative face-body pixels, but also loses the activations of mouth.

Overall, AMP works stably on keeping all the negative and positive texture responses.
Moreover, it has ability to hold a good balance between objective and background.
This result concludes that AMP works better than both of AvgP and MP for aggregating features.
It's possible to use AMP as an alternative general operation for other tasks. 
For example, in the video classification, AMP can be used to aggregate/pool features along the temporal dimension.

\end{document}